%% file: ICVGIP-Latex-Template.tex
\documentclass[sigconf, review=false]{acmart}
\usepackage{hyperref}
\usepackage{hyperxmp}

\usepackage{booktabs} % For formal tables

\usepackage{pgfplots}
\pgfplotsset{compat=1.18}
\usepackage{tikz}
\usepackage{multirow}
\usepackage{subcaption}

\setcopyright{rightsretained}
\acmConference[ICVGIP'25]{16th Indian Conference on Computer Vision, Graphics and Image Processing}{December 2025}{Mandi, India}
\acmYear{2025}
\copyrightyear{2025}

\acmPrice{15.00}

\begin{document}
\title{Are Detectors Fair to Indian IP-AIGC? A Cross-Generator Study}
\titlenote{Produces the permission block, and
  copyright information}

\author{Vishal Dubey}
\authornote{Lead author.}
\affiliation{%
  \city{Hyderabad}
  \country{India}
}
\email{vishaldubey0026@gmail.com}

\author{Pallavi Tyagi}
\authornote{Data curation; prompt engineering for the IP\mbox{-}AIGC dataset.}
\affiliation{%
  \city{Bengaluru}
  \country{India}
}
\email{tyagipallavi093@gmail.com}

\renewcommand{\shortauthors}{}

\begin{abstract}
Modern image editors can produce identity-preserving AIGC (IP-AIGC), where the same person appears with new attire, background, or lighting. The robustness and fairness of current detectors in this regime remain unclear, especially for under-represented populations. We present what we believe is the first systematic study of IP-AIGC detection for Indian and South-Asian faces, quantifying cross-generator generalization and intra-population performance. We assemble Indian-focused training splits from FairFD and HAV-DF, and construct two held-out IP-AIGC test sets (HIDF-img-ip-genai and HIDF-vid-ip-genai) using commercial web-UI generators (Gemini and ChatGPT) with identity-preserving prompts. We evaluate two state-of-the-art detectors (AIDE and Effort) under pretrained (PT) and fine-tuned (FT) regimes and report AUC, AP, EER, and accuracy. Fine-tuning yields strong in-domain gains (for example, Effort AUC 0.739 to 0.944 on HAV-DF-test; AIDE EER 0.484 to 0.259), but consistently degrades performance on held-out IP-AIGC for Indian cohorts (for example, AIDE AUC 0.923 to 0.563 on HIDF-img-ip-genai; Effort 0.740 to 0.533), which indicates overfitting to training-generator cues. On non-IP HIDF images, PT performance remains high, which suggests a specific brittleness to identity-preserving edits rather than a generic distribution shift. Our study establishes IP-AIGC-Indian as a challenging and practically relevant scenario and motivates representation-preserving adaptation and India-aware benchmark curation to close generalization gaps in AIGC detection.\footnote{This Work is independent research.} Code link is also shared.
\footnote{https://github.com/vishal-dubey-0026/DF\textunderscore{}India}

\end{abstract}




\maketitle

\input{samplebody-conf}

\bibliographystyle{ACM-Reference-Format}
\bibliography{ICVGIP-Latex-Template}

\end{document}

%% file: samplebody-conf.tex
\section{Introduction}
Reliable detection of AI-generated content (AIGC) is increasingly critical as modern generators produce photo-realistic media that challenge both humans and automated detectors \cite{kang2025hidf,wang2025spotting}. While progress in large-scale benchmarks and baselines has improved overall accuracy \cite{wang2025spotting,DeepfakeBench_YAN_NEURIPS2023,krubha2025robust,wang2019racial}, recent work highlights sharp failures on unseen generators, class imbalance, and minority populations \cite{wu2025preserving,lin2024preserving,ju2024improving,krubha2025robust,xu2024analyzing}. This raises urgent questions about fairness and population coverage, especially for groups under-represented in public datasets.

A growing line of work measures demographic bias in forgery detection across race/skin-tone, gender, and age \cite{liu2025thinking,lin2025ai}. Indian-centric resources \cite{narayan2023df,singh2019disguised,deo2025indicsideface,kuckreja2024indiface} further reveal domain shifts specific to Indian subjects and scenes. However, the identity-preserving AIGC (IP-AIGC) setting edits \cite{khan2025instaface} that keep the same person while altering attire/background/lighting, remains under-tested for South Asian populations, and detectors trained on standard corpora often underperform here \cite{kang2025hidf,yansanity}.

This paper studies generalization and fairness for Indian/South-Asian faces under IP-AIGC. We ask: Do state-of-the-art detectors degrade on identity-preserving generations with Indian-specific prompts and contexts? We combine Indian-focused training splits with two representative detectors: AIDE \cite{yansanity} and Effort \cite{yanorthogonal}, and evaluate on held-out, web-UI generators (ChatGPT, Gemini) for IP-AIGC–Indian. We report ACC/AUC/AP/EER.

\begin{table}[h]
\centering
\caption{Training and testing datasets}
\label{tab:dataset-info}
\scalebox{0.85}{%  <-- 0.85–0.95 as needed
\begin{tabular}{@{}lcc@{}}
\toprule
Dataset (no. of images) & Real & Fake \\
\midrule
FairFD            & 10308 & 61678   \\
HAV-DF-train            & 2444 & 3759   \\
HAV-DF-test            & 535 & 931   \\
HIDF-img         & 747 & 747  \\
HIDF-vid & 142 & 223  \\
HIDF-img-ip-genai & 747 & 535 \\
HIDF-vid-ip-genai & 142 & 108 \\
\bottomrule
\end{tabular}%
}
\end{table}

\begin{figure}[th]
\begin{center}
\includegraphics[width=1.0\columnwidth]{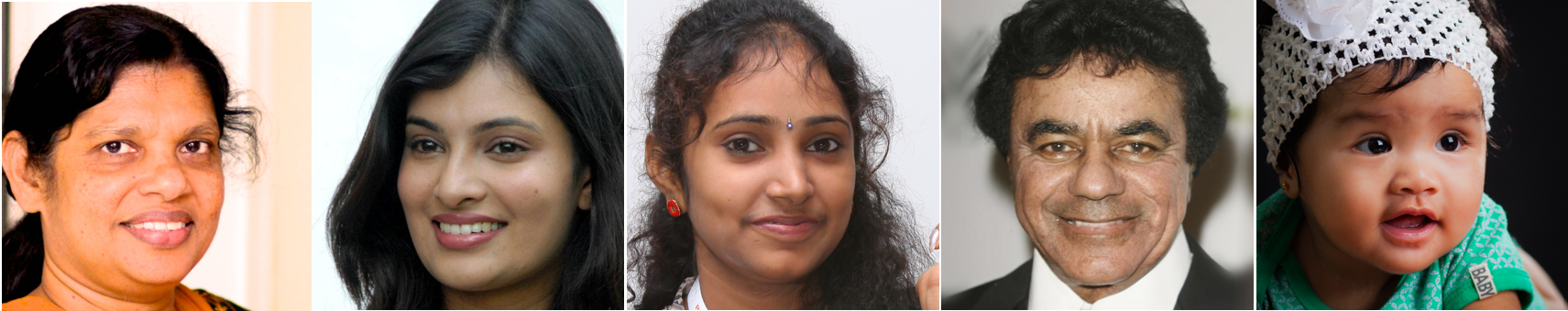}\\
\includegraphics[width=1.0\columnwidth]{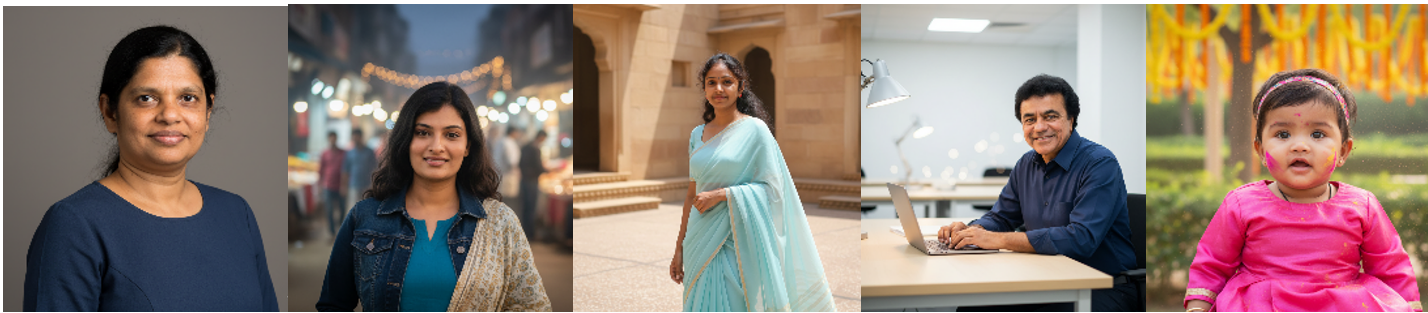}

\caption[width=0.5\columnwidth]{Samples of HIDF-img (real, 1st row) and HIDF-img-ip-genai(fake, 2nd row)}
\label{hidf_img}
\end{center}
\end{figure}

\begin{figure}[h]
\begin{center}
\includegraphics[width=1.0\columnwidth]{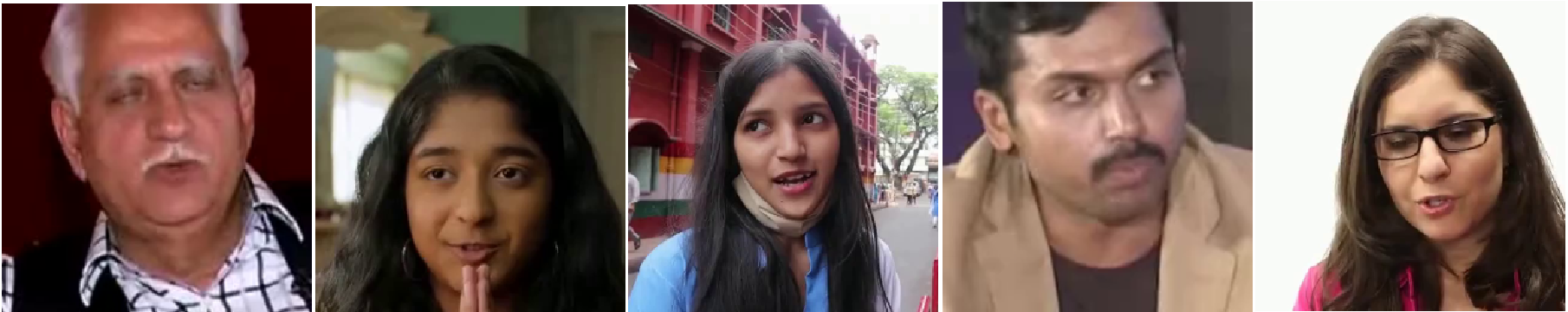}\\
\includegraphics[width=1.0\columnwidth]{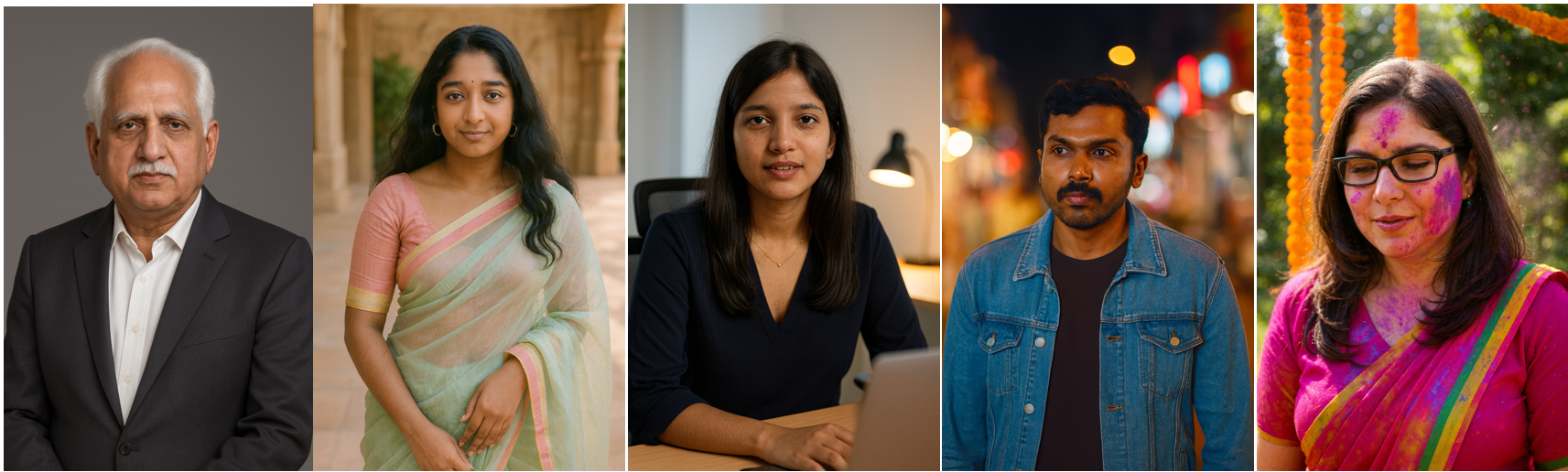}

\caption[width=0.5\columnwidth]{Samples of HIDF-vid (real, 1st row) and HIDF-vid-ip-genai(fake, 2nd row)}
\label{hidf_vid}
\end{center}
\end{figure}

\begin{table}[t]
\centering
\caption{Eval. results of AIDE}
\label{tab:model-accuracy-aide}
\resizebox{\columnwidth}{!}{%
\begin{tabular}{@{}lcccc@{}}
\toprule
Testset (PT/FT) & Acc $\uparrow$ & AP $\uparrow$ & AUC $\uparrow$ & EER $\downarrow$ \\
\midrule
HAV-DF-test            & 0.577/0.752 & 0.684/0.879 & 0.535/0.809 & 0.484/0.259  \\
HIDF-img         & 0.429/0.568 & 0.432/0.659 & 0.426/0.648 & 0.559/0.394 \\
HIDF-vid & 0.473/0.665 & 0.636/0.686 & 0.527/0.608 & 0.450/0.450  \\
HIDF-img-ip-genai & 0.804/0.555 & 0.842/0.413 & 0.923/0.563 & 0.123/0.461  \\
HIDF-vid-ip-genai & 0.544/0.532 & 0.587/0.458 & 0.593/0.570 & 0.436/0.464  \\
\bottomrule
\end{tabular}%
}
\end{table}

\begin{table}[t]
\centering
\caption{Eval. results of Effort}
\label{tab:model-accuracy-effort}
\resizebox{\columnwidth}{!}{%
\begin{tabular}{@{}lcccc@{}}
\toprule
Testset (PT/FT) & Acc $\uparrow$ & AP $\uparrow$ & AUC $\uparrow$ & EER $\downarrow$ \\
\midrule
HAV-DF-test            & 0.664/0.867 & 0.864/0.971 & 0.739/0.944 & 0.353/0.125  \\
HIDF-img         & 0.874/0.899 & 0.985/0.973 & 0.986/0.974 & 0.042/0.066 \\
HIDF-vid & 0.756/0.758 & 0.904/0.887 & 0.851/0.839 & 0.225/0.225  \\
HIDF-img-ip-genai & 0.670/0.497 & 0.605/0.398 & 0.740/0.533 & 0.321/0.447  \\
HIDF-vid-ip-genai & 0.524/0.348 & 0.448/0.302 & 0.535/0.249 & 0.471/0.676  \\
\bottomrule
\end{tabular}%
}
\end{table}

\section{Methodology}
\textbf{Data.} Training uses the Indian split from FairFD \cite{liu2025thinking} and the train split of HAV-DF \cite{kaur2024hindi}. Testing uses HAV-DF-test and an Indian-only subset of \cite{kang2025hidf}, yielding HIDF-img and HIDF-vid (video frames subsampled). We further construct IP-AIGC test sets (HIDF-img-ip-genai and HIDF-vid-ip-genai) using Gemini-2.5-Flash and ChatGPT image editing via their web UIs with identity-preserving prompts (Figure~\ref{hidf_img}--\ref{hidf_vid}). Dataset sizes appear in Table~\ref{tab:dataset-info}.

\textbf{Sample prompts.}
{\small
\begin{enumerate}
  \item \texttt{P1: Generate the **same subject** during a **daytime festive event**; outfit: **bright saree or salwar** with **subtle Holi-like color traces** on clothing and **only a light dusting on cheeks/forehead** (do **not** obscure key facial landmarks). Background: **outdoor garden with marigold garlands**. Soft backlight for subject separation; **fast shutter feel** to freeze stray powder. Identity, age, hair texture, and skin tone must match the reference.}
  \item \texttt{P2: Use the provided reference image of the **same Indian male**. **Preserve facial identity, bone structure, and skin tone**; no age change or facial slimming. Generate a waist-up **formal studio portrait** in a **charcoal business suit** with white shirt, no tie. **Neutral gray seamless background**, **three-point Lighting** (key at 45°, soft rim, subtle fill), **50mm equivalent, f/2.8**. Natural skin texture (no plastic skin), clean background, no text.}
\end{enumerate}
}

\textbf{Models and training.} We evaluate AIDE and Effort. For AIDE, we fine-tune (FT) from the GenImage pretrained (PT) weights; for Effort, we use the default PT weights and fine-tune to convergence. Training runs on a single Colab L4 (22.5\,GB). Effort: batch 40, lr $2\times10^{-4}$; AIDE: batch 32, lr $1\times10^{-6}$. Evaluation metrics include Accuracy (Acc), Average Precision (AP), area under ROC (AUC), Equal Error Rate (EER).

\section{Results}
Tables~\ref{tab:model-accuracy-aide}--\ref{tab:model-accuracy-effort} summarize frame/video performance for PT vs.\ FT.

\paragraph{In-domain improvements.} Fine-tuning substantially improves in-domain generalization on HAV-DF-test for both methods. AIDE AUC rises from 0.535 (PT) to 0.809 (FT) with EER dropping from 0.484 to 0.259; Effort AUC rises from 0.739 to 0.944 with EER from 0.353 to 0.125.

\paragraph{ID-preserving, held-out generators (IP-AIGC-Indian).} On HIDF-img-ip-genai, both methods lose cross-generator robustness after FT: AIDE AUC falls from 0.923 (PT) to 0.563 (FT) with EER worsening from 0.123 to 0.461; Effort AUC drops from 0.740 to 0.533 (EER 0.321 to 0.447). Video results are similar or worse: on HIDF-vid-ip-genai, Effort AUC declines from 0.535 to 0.249 and AIDE shows a mild decline (0.593 to 0.570) alongside higher EER (0.436 to 0.464).

\paragraph{Non-IP HIDF} On HIDF-img (non-IP), Effort PT is already strong (AUC 0.986) and remains competitive after FT (0.974); AIDE improves from 0.426 to 0.648. This contrast with IP-AIGC suggests adaptation can help when the target distribution is similar to training, but may harm identity-preserving, held-out edits.

\paragraph{Additional evidence on IP-AIGC drops.} The degradation on IP-AIGC is also visible in AP and ACC. For AIDE on HIDF-img-ip-genai, AP drops from 0.842 (PT) to 0.413 (FT). Effort shows a similar decline from 0.605 to 0.398. On HIDF-vid-ip-genai, AIDE AP decreases from 0.587 to 0.458, and Effort from 0.448 to 0.302, with corresponding accuracy reductions. These consistent declines indicate that FT overfits to training-generator cues and transfers poorly to identity-preserving edits.

\paragraph{Takeaway.} FT delivers clear in-domain gains yet induces notable overfitting to training-generator cues, hurting identity-preserving cross-generator generalization in Indian cohorts.

\section{Conclusion}
We present an IP-AIGC–Indian evaluation showing that fine-tuning strong detectors increases in-domain accuracy but degrades robustness on held-out, identity-preserving edits (image and video). These results indicate that present SOTA remains brittle in Indian IP-AIGC settings. Future work will target identity-aware adaptation (e.g., residual-subspace tuning with fairness constraints) and controlled expression-only edits to further stress generalization.